\documentclass[runningheads]{llncs}

\usepackage[T1]{fontenc}
\usepackage{graphicx}

\usepackage{amsmath}
\usepackage{amsfonts}
\usepackage{amssymb}
\usepackage{booktabs}
\usepackage{pdflscape}
\usepackage{bbm}
\usepackage{hyperref}

\usepackage{rotating}

\begin{document}

\title{Flexible Counterfactual Explanations \\ with Generative Models}

\titlerunning{Abbreviated paper title}

\author{Stig Hellemans\inst{1,2}\orcidID{0000-0001-9441-3882} \and
Andres Algaba\inst{1}\orcidID{0000-0002-0532-3066} \and
Sam Verboven\inst{1}\orcidID{0000-0002-1742-5561} \and
Vincent Ginis\inst{1,3}\orcidID{0000-0003-0063-9608}}

\titlerunning{ }
\authorrunning{ }

\institute{
Vrije Universiteit Brussel \and
Universiteit Antwerpen \and
Harvard University \\
\email{stig.hellemans@student.uantwerpen.be} \quad
\email{andres.algaba@vub.be} \quad
\email{sam.verboven@vub.be} \quad
\email{ginis@seas.harvard.edu}
}
\maketitle              

\renewcommand{\thefootnote}{*}

\begin{abstract}
Counterfactual explanations provide actionable insights to achieve desired outcomes by suggesting minimal changes to input features. However, existing methods rely on fixed sets of mutable features, which makes counterfactual explanations inflexible for users with heterogeneous real-world constraints. Here, we introduce Flexible Counterfactual Explanations, a framework incorporating counterfactual templates, which allows users to dynamically specify mutable features at inference time. In our implementation, we use Generative Adversarial Networks (FCEGAN), which align explanations with user-defined constraints without requiring model retraining or additional optimization. Furthermore, FCEGAN is designed for black-box scenarios, leveraging historical prediction datasets to generate explanations without direct access to model internals. Experiments across economic and healthcare datasets demonstrate that FCEGAN significantly improves counterfactual explanations' validity compared to traditional benchmark methods. By integrating user-driven flexibility and black-box compatibility, counterfactual templates support personalized explanations tailored to user constraints.\footnote{Code available at \url{https://github.com/stighellemans/fcegan}}


\end{abstract}

\begin{figure}[t!]
    \centering
    \includegraphics[width=0.95\linewidth]{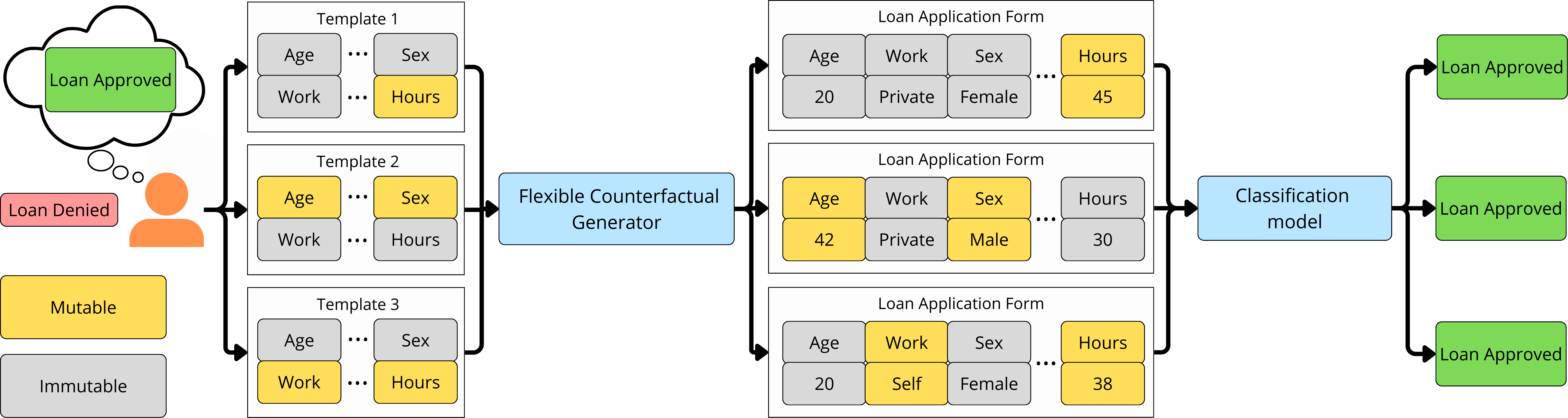}
    \caption{\textbf{Flexible Counterfactual Explanations GAN (FCEGAN)}. This diagram illustrates an individual denied a loan by a classification model based on input features such as age, work type, and sex. Users receiving an unfavorable prediction can explore which changes to their features might result in a favorable outcome. Unlike existing methods, FCEGAN empowers users to specify which features are mutable via a \textit{counterfactual template}, providing flexibility in determining actionable changes. The model generates counterfactual explanations based on user preferences without requiring retraining, allowing individuals to identify and experiment with the most suitable feature modifications to achieve their desired prediction.}
    \label{fig:overview framework}
\end{figure}

\section{Introduction}
\label{sec:intro}
In decision-making systems—such as loan approvals in banks or disease risk predictions in healthcare—users often seek actionable insights to achieve favorable outcomes \cite{tsirtsis2020decisions,grath2018interpretable,verma2020counterfactual,verma2024counterfactual}. For instance, a rejected loan applicant may want to know if increasing their income or reducing outstanding debts would result in loan approval. Similarly, patients classified as high-risk for developing a disease may wish to understand what changes in their lifestyle or medical parameters could lower their risk. Counterfactual explanations address this need by providing the smallest changes to input features that alter a model's prediction to a desired outcome \cite{degrave_2021_ai,mertes_2022_ganterfactual}.

A high-quality counterfactual explanation must satisfy several criteria \cite{wachter_2017_counterfactual}. It should be \textit{actionable}, meaning the proposed changes are feasible and only involve mutable features (e.g., income or exercise levels, but not age or genetic factors)~\cite{poyiadzi2020face,ustun2019actionable}. Explanations should also be \textit{realistic}, ensuring that they align with the data manifold \cite{joshi2019towards,pawelczyk2020learning}, and \textit{diverse}, offering multiple possible solutions for users to explore \cite{mothilal2020explaining}. Additional desiderata such as \textit{sparsity} (fewer changes) and \textit{proximity} (minimal distance to the original input) further enhance the actionability of counterfactual explanations \cite{karimi2021algorithmic}.

Despite their potential, existing methods for generating counterfactuals face key challenges. State-of-the-art counterfactual explanation generation methods— whether gradient optimization-based \cite{poyiadzi2020face,joshi2019towards,mothilal2020explaining,van2021interpretable} or generative \cite{nemirovsky_2022_countergan,vanlooveren_2021_conditional,singla_2021_explaining}—assume a fixed set of mutable features, whereas more flexible non-gradient optimization-based methods, such as constrained satisfaction problem solvers (SATs), fail to accommodate for traditional desiderata in counterfactual explanations, such as incorporating a realism loss \cite{barzekar2023achievable,rasouli2024care,salimi2023towards}. All these methods thus fail to account for users' diverse preferences and real-world constraints. For instance, increasing one's income may be feasible for one user but unrealistic for another. Another example is changing one's residential location: while relocating might be a plausible option for a single individual, it could be impractical for someone with family commitments or legal residency constraints. Moreover, existing approaches often require retraining or regenerating explanations when users modify their preferences, reducing their practical usability. 

Another significant limitation is the lack of robust solutions for black-box settings, where access to the model's internal parameters is restricted, such as with proprietary systems or federated learning environments. Many existing techniques assume direct access to or differentiability of the underlying model, which becomes infeasible in these scenarios \cite{joshi2019towards,mothilal2020explaining,vanlooveren_2021_conditional,karimi2020model}.

To address these limitations, we propose Flexible Counterfactual Explanations using Generative Adversarial Networks (FCEGAN) (Fig. \ref{fig:overview framework}). FCEGAN introduces the novel concept of counterfactual templates, which encode information about which features are allowed to change and which must remain fixed. This approach empowers users to dynamically specify mutable features, offering unprecedented flexibility in exploring actionable changes. Beyond generative models, counterfactual templates also enhance gradient-based optimization methods by embedding these templates directly into the optimization process. This ensures that only the user-defined mutable features are modified, significantly improving the validity and alignment of counterfactual explanations with user constraints.

Additionally, FCEGAN supports black-box scenarios by utilizing historical predictions of the model rather than requiring direct access to its internals. By combining the flexibility of user-driven templates with robust gradient-based optimization and black-box compatibility, FCEGAN provides a versatile, practical, and scalable solution for generating actionable, realistic, and personalized counterfactual explanations.

In summary, our approach addresses two critical limitations of existing counterfactual generation methods: (1) enhancing adaptability to user preferences through the introduction of counterfactual templates, which allow users to dynamically specify mutable features, and (2) enabling counterfactual explanations in black-box scenarios by leveraging historical model predictions without requiring access to internal model parameters. By integrating these advancements, FCEGAN and the template-enhanced gradient-based optimization methods provide a flexible framework for generating realistic, actionable counterfactuals, making them highly applicable to domains like healthcare and finance, where user constraints and interpretability are crucial.

\section{Background and Related Work}
Counterfactual explanations can be generated using a variety of methods, which are broadly classified into optimization-based and generative approaches. Op-timization-based methods aim to optimize an objective function subject to certain constraints \cite{poyiadzi2020face,ustun2019actionable,joshi2019towards,mothilal2020explaining,karimi2020model}. These methods focus on minimizing a distance function, with one key constraint ensuring that the resulting counterfactual satisfies the desired class prediction. Within this category, optimization techniques can be further divided into gradient-based and non-gradient-based approaches. Discrete non-gradient methods, such as constraint satisfaction problem solvers (SATs), form a subclass that often fail to account for traditional desiderata in counterfactual explanations, such as incorporating a realism loss \cite{barzekar2023achievable,rasouli2024care,salimi2023towards}.

 An advantage of optimization-based methods is having great flexibility in constructing custom distance functions and additional constraints. However, this flexibility also results in the need to construct a new optimization search space for each task. It is often unclear which distance function is optimal since the difficulty of adjusting a feature is a personal matter. Furthermore, these methods may lack assurance that the generated explanations are realistic. To address this issue, reconstruction loss from auto-encoders or outputs from GAN discriminators can be incorporated as additional constraints in gradient-based methods \cite{joshi2019towards,van2021interpretable}. 
 
 Despite their benefits, these methods suffer from slow generation speeds, as multiple optimization steps are required to generate each candidate explanation. To accelerate the process, techniques such as leveraging class prototypes—abstract representations of samples belonging to a specific class in the latent space—can guide the optimization path more effectively \cite{van2021interpretable}. Yang et al.~\cite{yang2022mace} achieve a speedup by identifying the most relevant features, focusing on the nearest samples that match the desired target class.

Generative models, such as GANs \cite{nemirovsky_2022_countergan,vanlooveren_2021_conditional,singla_2021_explaining}, inherently adhere to the data manifold and offer rapid inference speeds due to their one-shot generation capability, as they require training only once for the entire data space. However, a key limitation of existing GAN approaches is their assumption of a fixed set of mutable features, necessitating retraining when a different combination of features is considered. This restricts user flexibility in choosing which features to modify and which to keep immutable. Nemirovsky et al.~\cite{nemirovsky_2022_countergan} tackled this issue by introducing a method to respect immutable features during counterfactual generation. Specifically, their approach searches for counterfactual examples while automatically reverting any changes made to features marked as immutable, ensuring these features remain unaltered. However, the set of immutable features remains fixed, and users are still unable to dynamically select which features to keep unaltered. 

FCEGAN addresses this limitation by introducing a \textit{counterfactual template} that allows users to specify which features are mutable. This template serves as an additional input to the counterfactual generator, enabling users to dynamically control which features can be altered. FCEGAN incorporates techniques from the Conditional Tabular GAN (CTGAN) \cite{xu_modeling_2019}, including training-by-sampling \cite{engelmann_conditional_2021,zhao_ctab-gan_2021}, and the Wasserstein GAN \cite{arjovsky_towards_2017,arjovsky_wasserstein_2017} with gradient penalty (WGAN-GP) \cite{gulrajani_improvedwgan_2017}, to generate diverse and realistic synthetic tabular data \cite{xu_modeling_2019}. These methods effectively mitigate mode collapse \cite{srivastava2017veegan}, which can otherwise compromise the diversity of the generated counterfactual explanations.

In black-box scenarios, where the underlying model's architecture and parameters are inaccessible, a promising solution is to leverage a dataset of the model's historical predictions. These datasets inherently capture the relationships between specific input features and the corresponding outputs. For instance, Nemirovsky et al.\cite{nemirovsky_2022_countergan} utilize this idea by weighting samples based on the model's previously assigned scores. This strategy enables the generation of counterfactuals that align with the model's prior behavior, even in the absence of direct access to its internal workings.

\begin{figure}[t!]
    \centering
    \includegraphics[width=0.95\linewidth]{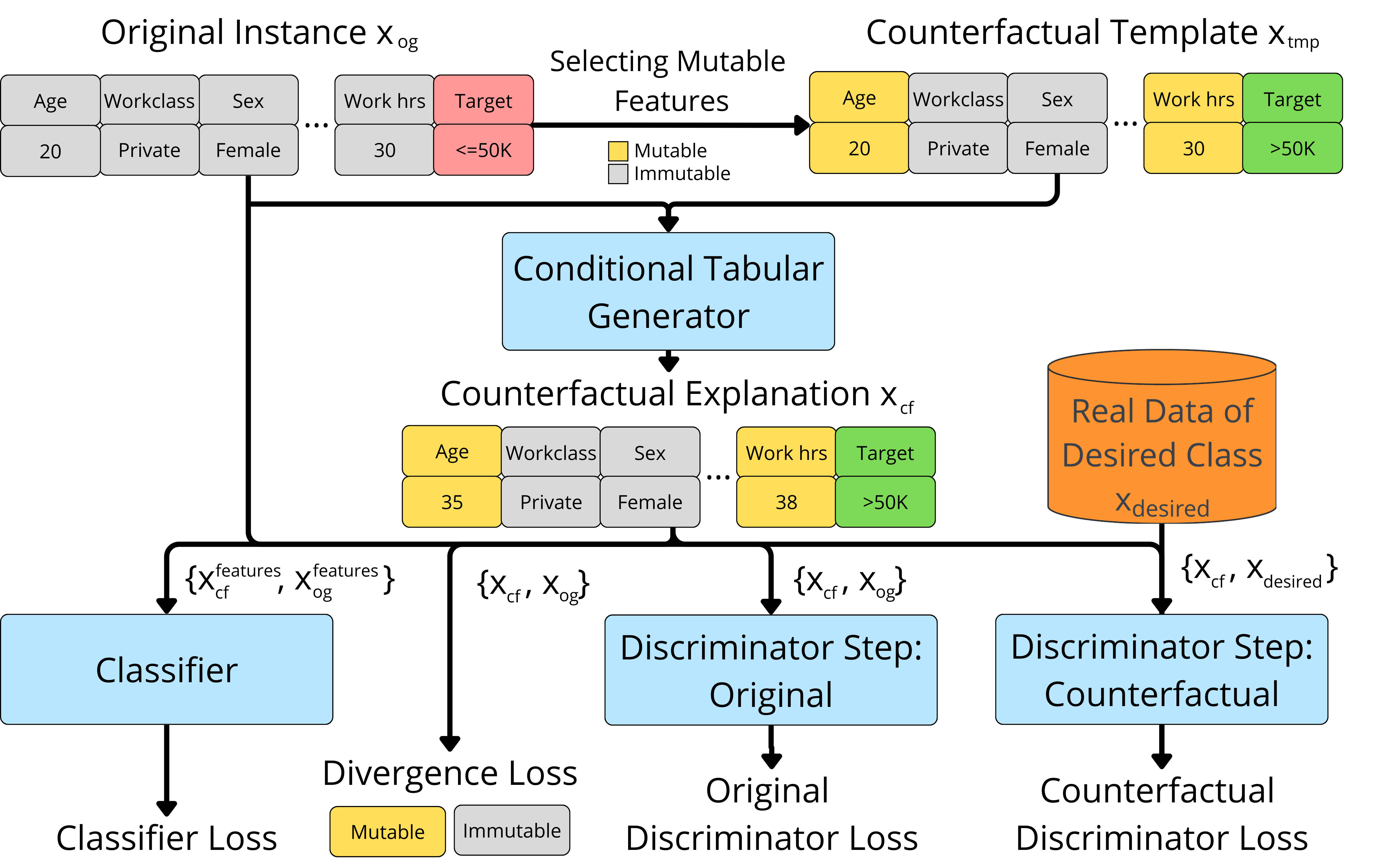}
    \caption{\textbf{FCEGAN Architecture}. The users' original features \(x_{og}\), concatenated with their (undesired) predictions, are input to the model. Normally the user would select which features are set mutable in the counterfactual template. During training, counterfactual templates \( x_{tmp} \) are generated by randomly setting a fraction of features as mutable, along with specifying a desired target \( y_{desired} \), which is especially important in multi-class settings. The original instance \( x_{og} \) and counterfactual template \( x_{tmp} \) are fed into the flexible counterfactual generator, which outputs the predicted counterfactuals \( x_{cf} \). To ensure realism, a combination of two discriminator losses is used: \( L_{D_{og}} \), comparing \( x_{cf} \) with the original instance \( x_{og} \), and \( L_{D_{cf}} \), comparing \( x_{cf} \) with real counterfactuals from the desired class \( x_{desired} \). To limit divergence, a divergence loss \( L_{div} \) is applied, computed separately for mutable and immutable features. An optional classifier loss \( L_{clas} \) can guide the generator to produce valid counterfactuals aligned with the desired target.}
    \label{fig:architecture}
\end{figure}

\section{Flexible Counterfactual Explanations}
\label{sec:methodology}

FCEGAN addresses two key limitations in existing counterfactual generation methods: (1) it allows users to dynamically specify which features are mutable through a \textit{counterfactual template}, and (2) it can support black-box models by relying solely on historical predictions rather than requiring direct access to the model. The counterfactual template empowers users to dynamically specify which features can be modified without retraining the model, as illustrated in Fig.~\ref{fig:architecture}. Furthermore, these templates can be integrated into a gradient optimization method to guide the generation process, ensuring targeted and adaptable counterfactual explanations.

\subsection{Specifying mutable features}
The core idea behind specifying mutable features lies in the use of the counterfactual template \( x_{tmp} \), which indicates which features can be modified. This template, along with the original instance \( x_{og} \) and the desired target label, serves as input to a conditional generator \cite{mirza_conditional_2014}. For multi-class classification tasks, specifying the desired target is especially important \cite{grandini2020metrics}. To encode mutable features, a copy of the original instance is created where the mutable features are masked, and the target label is replaced with the desired one. This template gives the generator clear guidance on which features it is allowed to change while ensuring the rest remain intact.

The counterfactual explanation \( x_{cf} \) is then generated by a conditional generator, implemented as part of a Wasserstein GAN \cite{arjovsky_towards_2017,arjovsky_wasserstein_2017} with gradient penalty (WGAN-GP) \cite{gulrajani_improvedwgan_2017}. Although the generator can perturb both mutable and immutable features, any changes made to immutable features are reset to their original values, as introduced by Nemirovsky et al.~\cite{nemirovsky_2022_countergan}. This ensures that the generated counterfactuals respect the immutability constraints.

During gradient-based optimization, the counterfactual template plays a key role by resetting immutable features to their original values from the template after each gradient step. This ensures that the optimization process focuses exclusively on modifying mutable features, adhering to the predefined constraints.

\subsection{FCEGAN training}
\label{subsec:fcegan-training}
To produce realistic and valid counterfactual explanations, we employ a combination of loss functions. First, two discriminator losses are used to promote realism. The loss \( L_{D_{og}} \) ensures that the generated counterfactual \( x_{cf} \) remains close to the original instance \( x_{og} \), while \( L_{D_{cf}} \) compares \( x_{cf} \) to real counterfactuals \( x_{desired} \) sampled from the target class. The real counterfactuals are selected using the training-by-sampling method described in CTGAN~\cite{xu_modeling_2019}.

To limit the extent to which the counterfactual deviates from the original instance, we introduce a divergence loss \( L_{div} \). This loss is applied specifically to mutable features, as our experiments demonstrated that enforcing mutability constraints (\( L_{m,div} \)) is sufficient (see Appendix Fig.~\ref{fig:divergence-tune}). Consequently, the divergence loss for immutable features (\( L_{i,div} \)) is omitted. The divergence loss is defined as:
\begin{equation}
\begin{aligned}
L_{div} = L_{m,div} = \frac{1}{\left|\mathbf{x}\right|} \lambda_{m} \cdot [d_{cont}(\mathbf{x}_{m,og}, \mathbf{x}_{m,cf}) + d_{cat}(\mathbf{x}_{m,og}, \mathbf{x}_{m,cf})] \quad (\lambda_{m} \in \mathbb{R}),
\end{aligned}
\label{eq:div-loss}
\end{equation}
where \( d_{cont} \) measures the distance between normalized continuous features using Mean Squared Error (MSE), and \( d_{cat} \) evaluates one-hot encoded categorical features using cross-entropy loss. The total distance is normalized by the number of features \( |\mathbf{x}| \).

Finally, a classifier loss \( L_{clas} \) ensures that the generated counterfactual achieves the desired target class. This loss is calculated as the cross-entropy between the prediction \( y_{cf} \) for the generated counterfactual and the specified target \( y_{desired} \).

By combining these losses, FCEGAN generates partially mutable counterfactual explanations that are both realistic and satisfy user-defined immutability constraints. To verify that the synthetic counterfactuals produce the desired predictions, they are re-evaluated by the classification model, and the fraction of valid counterfactuals is reported to the user.

\subsection{Enabling black-box training}
In black-box settings, models are assumed to be inaccessible during training. A solution in such scenarios is to leverage a dataset containing the model's historical predictions. These datasets inherently capture the relationships between input features and corresponding outputs.

Nemirovsky et al.~\cite{nemirovsky_2022_countergan} build on this idea by weighting the discriminator loss based on the model's prediction scores. Samples with high prediction scores— indicating that they contain features characteristic of the desired output—are given greater influence during generator training.

We adopt a similar approach but simplify it by including \textbf{only} samples in \( L_{D_{cf}} \) that are predicted as belonging to the desired class. The classifier loss \( L_{clas} \), which is unavailable in a black-box setting, can be omitted. This is because \( L_{D_{cf}} \) already provides sufficient information about which features are important for generating synthetic samples of the desired class. Our experiments confirm this hypothesis (see Section \ref{sec:results}).

\subsection{Quality measures}
\label{subsec:quality measures}
To evaluate the quality of counterfactual explanations in line with the desired desiderata outlined in the Introduction, we define several quality measures (Table \ref{tab:measures}). The \textit{valid counterfactual fraction} represents the proportion of counterfactuals that successfully predict the desired target class \(y_{desired}\). This metric is critical, as the validity of counterfactuals cannot be guaranteed after their initial generation. It serves as the most important evaluation measure, reflecting the method's ability to identify the characteristics required to produce valid counterfactual explanations. 

To assess categorical proximity or divergence, we use the \textit{mean fraction of categorical columns changed} \cite{mothilal2020explaining}, while for continuous features, the \textit{mean/max percentile shift}  is employed \cite{pawelczyk2020learning}. 

The \textit{fakeness/realism} of counterfactuals is measured using an independent CTGAN discriminator trained on the original training data. Higher fakeness scores indicate less realistic samples, but the scale of this measure is dataset-dependent and should always be contextualized by comparing it to the fakeness scores of real samples. Lastly, \textit{diversity} is quantified as the expected proximity or divergence between pairs of counterfactuals, providing a measure of variation across the generated explanations \cite{mothilal2020explaining}.

\clearpage

\begin{sidewaystable}[h!]
    \centering
    \renewcommand{\arraystretch}{1.3} 
    \setlength{\tabcolsep}{8pt} 
    \begin{tabular}{p{4cm} p{9cm}} 
        \toprule
        \textbf{Measure} & \textbf{Formula} \\ 
        \midrule
        $\uparrow$ Counterfactual prediction \label{meas: cf pred} & 
        $f\left(\textbf{x}^{features}_{cf}\right)$ \\ 
        
        $\uparrow$ Valid counterfactual fraction & 
        $\mathbb{E}_{x_{cf} \sim P(x_{cf})} \left[ \mathbbm{1}\left(\textbf{x}^{features}_{cf} = y_{desired}\right) \right]$ \\ 
        
        $\downarrow$ Categories changed \label{meas: cat changed} \cite{mothilal2020explaining} & 
        $\frac{1}{\left|\mathbf{c}\right|}\sum_{i=1}^{\left|\mathbf{c}\right|}\mathbbm{1}\left(c_{i,cf} \neq c_{i, og}\right)$ \\ 
        
        $\downarrow$ Mean percentile shift \label{meas: mean perc shift} \cite{pawelczyk2020learning} & 
        $\frac{1}{\left|\mathbf{u}\right|}\sum_{i=1}^{\left|\mathbf{u}\right|}\left(cdf\left(u_{i, cf}\right) - cdf\left(u_{i, og}\right)\right)$ \\ 
        
        $\downarrow$ Max percentile shift \label{meas: max perc shift} \cite{pawelczyk2020learning} & 
        $\max_{i}\left(cdf\left(u_{i, cf}\right) - cdf\left(u_{i, og}\right)\right)$ \\ 
        
        $\downarrow$ Fakeness \label{meas: fakeness} & 
        $D_{CTGAN}\left(\mathbf{x}_{cf}\right)$ \\ 
        
        $\uparrow$ Categorical diversity \label{meas: cat div} \cite{mothilal2020explaining} & 
        $\mathbb{E}_{x^{i}_{cf}, x^{j}_{cf} \sim P(x_{cf})} \left[ \text{cat\_changed}(x^{i}_{cf}, x^{j}_{cf}) \mid x^{i}_{cf} \neq x^{j}_{cf} \right]$ \\ 
        
        $\uparrow$ Continuous diversity \label{meas: cont div} \cite{mothilal2020explaining} & 
        $\mathbb{E}_{x^{i}_{cf}, x^{j}_{cf} \sim P(x_{cf})} \left[ \text{mean\_percentile\_shift}(x^{i}_{cf}, x^{j}_{cf}) \mid x^{i}_{cf} \neq x^{j}_{cf} \right]$ \\ 
        \bottomrule
    \end{tabular}
    \caption{Quality measures of Counterfactual Explanations. 
    $cdf(\cdot)$ - Cumulative Density Function; 
    $f(\cdot)$ - Classifier; 
    $D_{CTGAN}\left(\cdot\right)$ - Independent CTGAN Discriminator.  The arrows indicate the desired direction of change for a given measure (e.g., a higher value being desirable is represented by $\uparrow$).} 
    \label{tab:measures}
\end{sidewaystable}

\clearpage

\section{Experiments}
\label{sec:results}
In this section, we present the four data sets that we use in our experimental setup. Then, we discuss the state-of-the-art benchmark methods and discuss the results. Finally, we highlight the usefulness of our approach in a black-box setting.

\subsection{Datasets}
To demonstrate the performance of FCEGAN and template-guided optimization, we evaluate these methods on four diverse datasets, chosen to highlight scenarios where counterfactual explanations could be particularly useful in altering predictions toward more desirable outcomes. Two medical datasets focus on predicting the future development of diseases: the \textit{Heart Disease Risk Prediction} dataset~\cite{lupague2023integrated} and the \textit{Diabetes Health Indicators} dataset, a cleaned version of the \textit{Behavioral Risk Factor Surveillance System 2015 (BRFSS-2015)}~\cite{cdc2015brfss}. In these cases, counterfactual explanations can inform patients about potential lifestyle changes required to influence disease progression. For instance, in the diabetes dataset, which contains three labels (\textit{No Diabetes, Pre-Diabetes, Diabetes}), counterfactual explanations enable experimentation with various lifestyle factors, identified as mutable in the counterfactual template, to evaluate their impact on health outcomes. 

In addition to medical use cases, we examine two business-oriented datasets: the \textit{Employee Attrition Classification} dataset and the \textit{Adult UCI Income} dataset, the latter representing a loan application setting. For employee attrition, counterfactual explanations can help organizations identify actionable changes to reduce the likelihood of employees leaving. Similarly, in loan applications, such explanations provide insights into factors influencing income classification, offering actionable strategies for improving outcomes. All datasets used in this study are publicly available on Kaggle. Missing data is omitted, and the data is split into train, validation, and test sets with a 60-20-20 ratio.

\begin{figure*}[t!]
    \centering
    \includegraphics[width=1\linewidth]{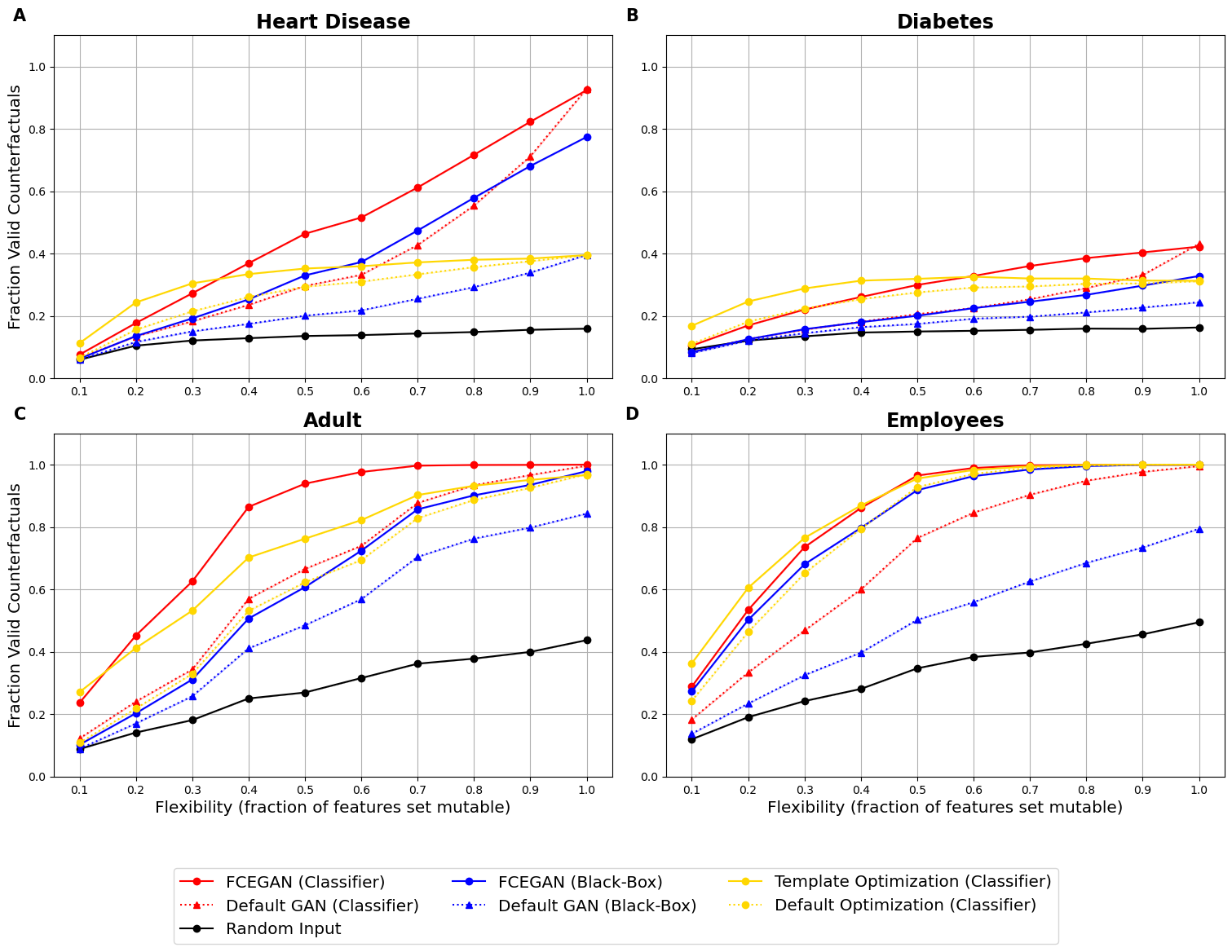}
    \caption{\textbf{Performance of Counterfactual Explanations with Increasing Flexibility.} As flexibility increases, a larger proportion of counterfactuals are valid, meaning they achieve the desired class prediction. This is crucial for improving search efficiency, as fewer explanations are discarded due to invalidity. Counterfactual templates substantially enhance this validity across all counterfactual methods compared to their default implementations. Another striking observation is that, when all features are set as mutable, the black-box FCEGAN significantly outperforms its default implementation, demonstrating its capability to enhance performance even in a non-flexible setting. All methods outperform random search (Random Input) by leveraging learned counterfactual characteristics, thereby improving search efficiency. No divergence constraints were applied in this comparison, as they were deemed unnecessary for the analysis. Other parameters: $\lambda_{clas} = (1/0)$, $\lambda_{D_{og}} = 0.5$, $\lambda_{D_{cf}} = 0.5$.}
    \label{fig:flexibility}
\end{figure*}

\begin{figure*}[t!]
    \centering
    \includegraphics[width=1\linewidth]{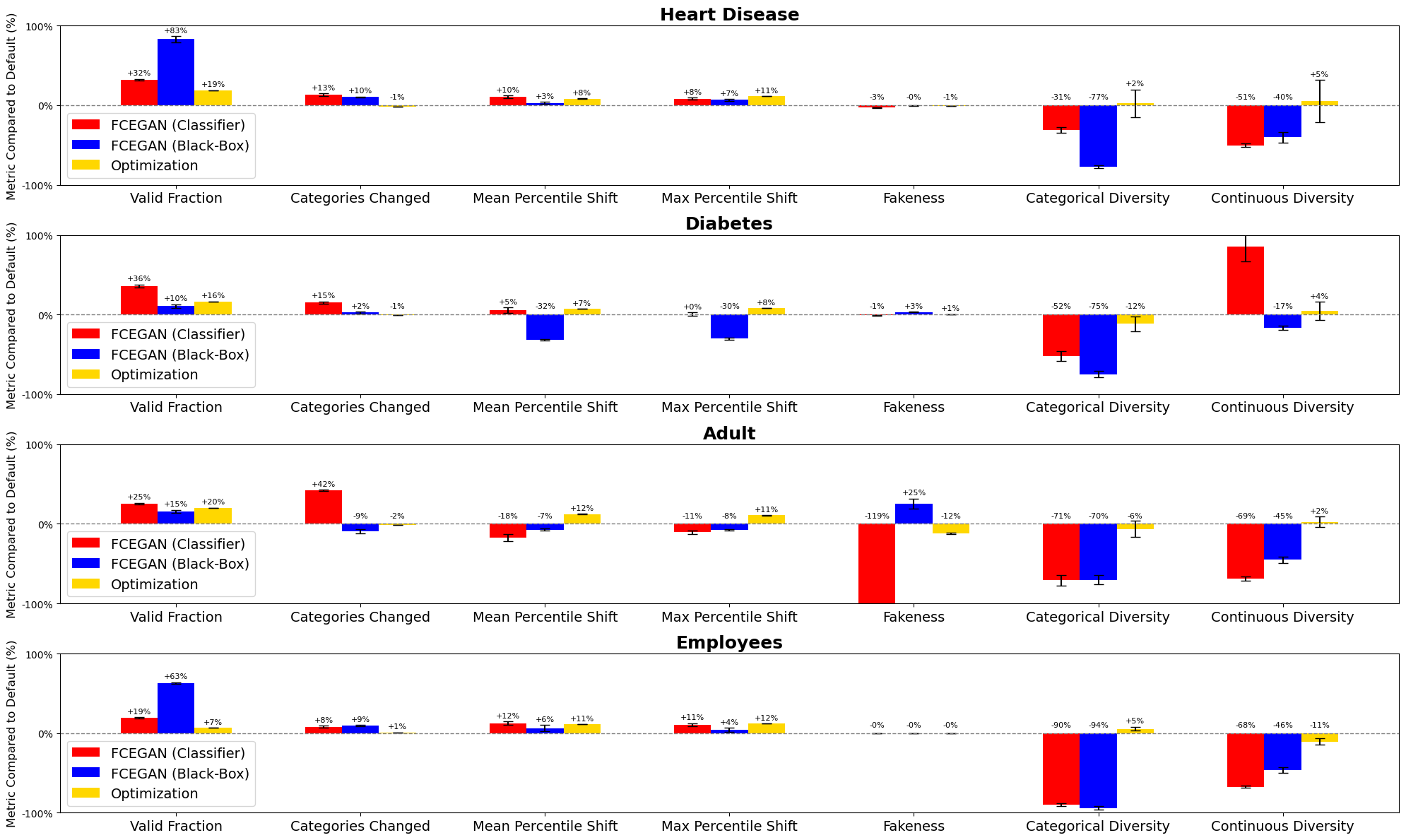}
\caption{\textbf{Impact of Counterfactual Templates on Counterfactual Quality.} The use of counterfactual templates notably increases the fraction of valid counterfactuals, as previously established. This figure further examines the associated changes in quality measures, as described in Section \ref{subsec:quality measures}. Both categorical divergence (changes in categories) and continuous divergence (mean/max percentile shifts) generally increase, though not uniformly. This increase in divergence arises because the model is constrained by immutable features, leading to greater variation in mutable features. The model effectively learns which features can be altered. Fakeness remains largely unchanged, except for certain methods applied to the Adult datasets, where it either increases or decreases depending on the approach. In contrast, counterfactual templates in FCEGAN models notably reduce diversity, which is an undesirable outcome.  However, enforcing stricter divergence constraints can help preserve diversity, as shown in Appendix Fig. \ref{fig:divergence-impact}. The presented measures represent the mean area under the curve (AUC) values of the flexibility graphs, as shown in Fig. \ref{fig:flexibility}, across five experiments. Error bars indicate the standard error of the mean (SEM). These values are normalized relative to their default implementations. No divergence constraints were applied during this comparison. $\lambda_{clas} = (1/0)$, $\lambda_{D_{og}} = 0.5$, $\lambda_{D_{cf}} = 0.5$.}
\label{fig:quality-metrics}
\end{figure*}

\subsection{Benchmarks}  
Since counterfactual templates represent a novel concept, no existing models can be directly compared to our FCEGAN and optimizer implementation. As a result, we benchmark against ablated methods that resemble prior counterfactual approaches. Specifically, our FCEGAN model, trained without counterfactual template knowledge or resetting immutable features, is comparable to the models proposed by Van Looveren (2021) \cite{vanlooveren_2021_conditional} and Nemirovsky (2022) \cite{nemirovsky_2022_countergan}. Both utilize a combination of classifier loss, divergence loss, and discriminator loss to train their counterfactual generators. In addition, we include a default gradient-based optimization method as an optimization benchmark, where no template guidance is applied during the optimization process.

In contrast, our method replaces the conventional L1 or L2 loss with a loss function specifically tailored for mixed tabular data (see Section \ref{subsec:fcegan-training}). Additionally, we employ a combination of two discriminator losses, unlike Nemirovsky (2022), who used a single discriminator loss, or Van Looveren (2021), who incorporated an auto-encoder. Detailed differences between these approaches can be found in their respective papers.

To establish a fundamental baseline, we also include random input generation, which requires no training or optimization. An improvement in the valid fraction achieved by our model over this baseline underscores its ability to learn the necessary characteristics for generating valid counterfactual explanations. Further details regarding the methods used are provided in Appendix Section \ref{subsec:architecture details}.

\subsection{Results}
As shown in Fig. \ref{fig:flexibility}, counterfactual templates significantly enhance the performance of counterfactual explanation methods. Both GAN-based and gradient-based optimization approaches benefit from these templates, achieving a higher fraction of valid counterfactuals—those that successfully meet the desired class prediction. This improvement in valid counterfactuals indicates greater search efficiency, as fewer explanations are discarded due to invalidity. Consequently, the methods demonstrate improved adaptability to constraints imposed by immutable features, effectively addressing the challenge of partial feature immutability. Feature-level examples are provided in Appendix Figures \ref{fig:whole dist} and \ref{fig:one sample dist}.

However, the advantage of counterfactual templates diminishes as the fraction of mutable features increases. With fewer constraints, the model requires less guidance from the template, reducing its overall utility.

And to enhance validity, methods guided by counterfactual templates generate counterfactuals of comparable quality but with reduced diversity in FCEGAN models (Fig. \ref{fig:quality-metrics}). Both categorical divergence (e.g., shifts in category distributions) and continuous divergence (e.g., changes in mean or maximum percentiles) generally increase, though not consistently. This variability stems from the model's ability to selectively modify mutable features while adhering to immutability constraints. However, imposing stricter divergence constraints can mitigate the reduction in diversity, as discussed in Appendix Fig.  \ref{fig:divergence-impact}.

\subsection{Black-box implementation}

Our FCEGAN architecture supports training in a black-box setting by omitting the classifier loss and utilizing a dataset of historical predictions from the classification model. This approach is particularly advantageous in scenarios where access to the model's internal parameters is restricted, such as in proprietary third-party systems or under federated and distributed learning protocols. As illustrated in Fig. \ref{fig:flexibility}, the black-box implementation of FCEGAN maintains strong performance, exhibiting only minor reductions in the fraction of valid counterfactuals on certain datasets, such as Employees, compared to classifier-based implementations. Furthermore, the black-box FCEGAN delivers quality metrics that are comparable to those achieved by its classifier-dependent counterparts.

\section{Conclusion}
\label{sec:conclusion}

Counterfactual explanations are a critical tool in decision-making systems, enabling users to understand and influence model predictions by identifying actionable changes to input features. Despite their utility, existing methods often lack the flexibility to accommodate user-specific constraints, rely on fixed sets of mutable features, and face challenges in black-box settings where model parameters are inaccessible. To address these issues, we proposed \textbf{Flexible Counterfactual Explanations using Generative Adversarial Networks (FCEGAN)}, which introduces counterfactual templates to allow users to dynamically specify mutable features. 

In addition to leveraging GAN-based approaches, our work enhances \textbf{gradient-based optimization methods} by incorporating counterfactual templates into the optimization process. These templates ensure that only mutable features are modified, improving the validity and alignment of counterfactual explanations with user-defined constraints. Tailored divergence losses further ensure the generated counterfactuals are realistic and actionable, making gradient-based optimization a viable and robust alternative for counterfactual generation.

Experimental results across diverse datasets show that both FCEGAN and template-guided gradient-based optimization methods significantly enhance the validity of counterfactual explanations, albeit at the expense of reduced diversity. This limitation can be partially mitigated through divergence constraints. Counterfactual templates enable users to personalize explanations without retraining models, while the framework's compatibility with black-box settings, using historical prediction datasets, ensures practicality in real-world scenarios.

A promising avenue for future research is to further enhance user agency by allowing manual adjustments to the counterfactual template. For instance, users could modify attributes such as weight or income and explore the resulting changes in other features needed to achieve the desired outcome. This interactive capability would empower users to test different scenarios and gain deeper insights into the relationships between features and predictions. Such advancements could expand the utility of counterfactual explanations in domains requiring high user engagement, such as healthcare and personalized financial planning.

By integrating user-driven flexibility, robust gradient-based optimization, and black-box compatibility, FCEGAN advances the state of the art in counterfactual explanation methods. Future research may also explore extending these methods to multi-modal datasets, incorporating temporal dynamics, and addressing fairness constraints, further broadening their applicability in critical decision-making domains.

\clearpage
\section*{Acknowledgements}
Andres Algaba acknowledges a fellowship from the Research Foundation Flanders under Grant No.1286924N. Vincent Ginis acknowledges support from Research Foundation Flanders under Grant No.G032822N and G0K9322N.

\bibliographystyle{unsrt}
\bibliography{biblio}

\setcounter{figure}{0}
\renewcommand{\thefigure}{A\arabic{figure}}

\setcounter{table}{0}
\renewcommand{\thetable}{A\arabic{table}}

\clearpage
\appendix
\section*{Appendix}
\label{sec:appendix}

We discuss our architecture details in Appendix \ref{subsec:architecture details} and show supplementary tables and figures. All experiments are conducted on a MacBook Air with an M1 chip and 8GB of RAM.

\setcounter{secnumdepth}{2}

\setcounter{table}{0}
\renewcommand{\thetable}{B\arabic{table}}

\setcounter{figure}{0}
\renewcommand{\thefigure}{C\arabic{figure}}

\section{Architecture Details}
\label{subsec:architecture details}
\paragraph{Classifier Configuration}
A neural network with two hidden layers is trained using the hyperparameters in Appendix Table \ref{tab:classifier}. To address class imbalance, a weighted loss function is employed.

\paragraph{Generator and Discriminator Configuration}
The generator and discriminator of the Conditional Tabular GAN (CTGAN) serve as the foundation for FCEGAN, including the independent CTGAN discriminator used for assessing fakeness \cite{xu_modeling_2019}. Detailed configuration settings are provided in Appendix Table \ref{tab:ctgan}. The generator leverages residual layers with concatenation to enhance feature learning, while the discriminator incorporates a PacGAN approach to improve training stability \cite{lin_pacgan_2018}. Additional details about the architecture are available at \url{https://github.com/sdv-dev/CTGAN}.

\begin{figure*}[t]
    \centering
    \includegraphics[width=0.8\linewidth]{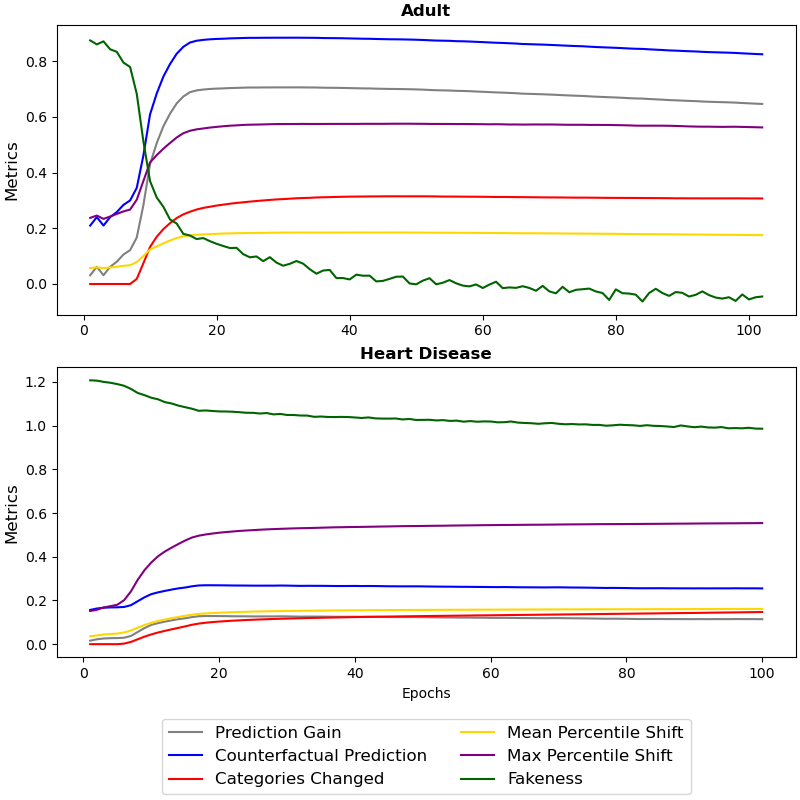}
    \caption{\textbf{Optimization performance.} Optimization saturates after 20-30 gradient steps in both datasets. Fakeness is measured with an independent CTGAN discriminator trained on the data distribution. The CTGAN data transformation is used, employing GMM-based normalization for continuous features.}
    \label{fig:optim finetune}
\end{figure*}

\paragraph{Optimization Configuration}
The loss for regularized gradient descent (RGD) consists of three components: (a) classifier loss $L_{clas}$, (b) divergence loss $L_{div}$ (Eq. \ref{eq:div-loss}), and (c) realism loss $L_{real}$, which is the output of a CTGAN discriminator trained on the data distribution. After each gradient step, immutable features are reverted back to their original values from the counterfactual template $x_{tmp}$. To accelerate the process, all samples requiring counterfactuals are batched, and a mean loss is optimized, significantly improving generation time per sample without compromising quality. Appendix Figure \ref{fig:optim finetune} shows that 20-30 gradient steps are sufficient for good results, after which training saturates. Hyperparameters are provided in Appendix Table \ref{tab:config-optimization}.

\begin{figure*}[t]
    \centering
    \includegraphics[width=1\linewidth]{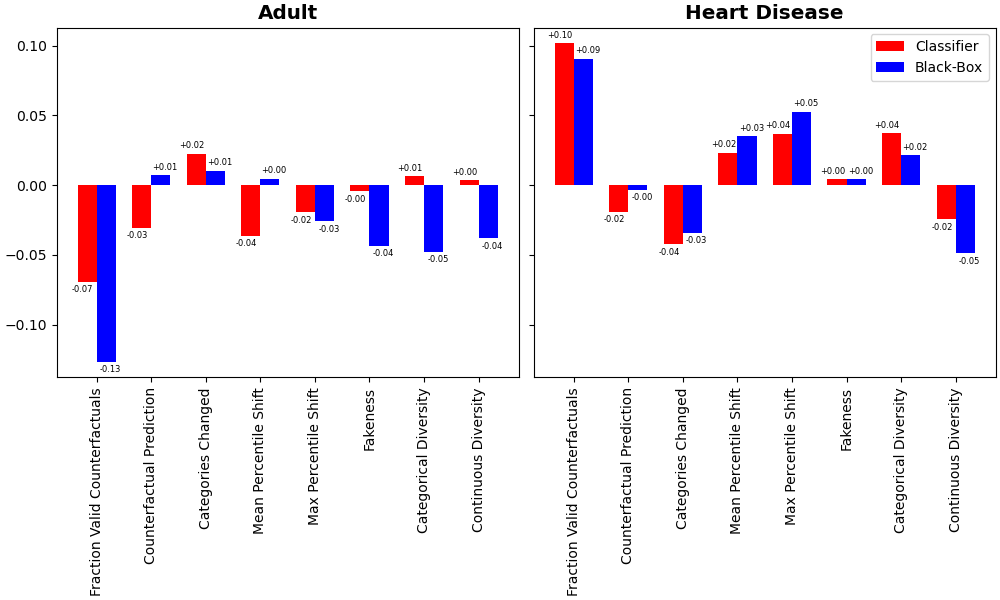}
   \caption{\textbf{Impact of Data Transformation on FCEGAN Performance (Relative to CTGAN Baseline).} This figure compares FCEGAN's counterfactual generation performance metrics using two data transformation approaches: a simple method involving standardization for continuous features and CTGAN's GMM-based normalization for continuous features. Results are shown for the Adult and Heart Disease datasets. The CTGAN method introduces stochastic mode selection, which results in non-deterministic classification predictions. In contrast, standardization avoids this variability and ensures reproducibility. No consistent performance advantage is observed for CTGAN-based transformations, with outcomes varying based on dataset characteristics. The simple standardization approach is therefore preferred for its reliability and consistency. Parameters:  $\lambda_{clas} = (1/0)$, $\lambda_{D_{og}} = 0.5$, $\lambda_{D_{cf}} = 0.5$.}
    \label{fig:no ctgan}
\end{figure*}

\paragraph{Data Transformation}  
To maintain consistency when including a classifier loss, FCEGAN relies on data transformation methods that align with those used in the classification model. In our study, we applied standardization for continuous features and one-hot encoding for categorical features. Alternatively, more complex methods, such as CTGAN \cite{xu_modeling_2019}, use a Gaussian Mixture Model (GMM) for normalizing continuous variables and one-hot encoding categorical features. CTGAN's mode-specific normalization produces both a one-hot encoded mode and a normalized value representing the position within the Gaussian mode. However, as illustrated in Fig. \ref{fig:no ctgan}, the choice of transformation method does not consistently influence counterfactual generation outcomes. While CTGAN's stochastic mode selection can lead to non-deterministic counterfactual classification predictions, standardization ensures reproducibility and avoids this issue, demonstrating reliable performance without systematic disadvantages.

\begin{table*}[h]
\centering
\renewcommand{\arraystretch}{1.3}
\setlength{\tabcolsep}{12pt}
\begin{tabular}{ll}
\hline
\textbf{Parameter}                      & \textbf{Value} \\ \hline
\texttt{batch\_size}                    & 300            \\
\texttt{hidden\_dimensions}             & [512, 512]     \\
\texttt{non\_linearity}                 & ReLU           \\
\texttt{adam\_betas}                    & [0.9, 0.999]   \\
\texttt{lr}                             & 0.001          \\
\texttt{epochs} (early stopping)                         & \{Adult/Employees: 10, Heart Disease/Diabetes: 50\}\\
\hline
\end{tabular}
\caption{\textbf{Classifier Configuration.} Longer training on the Heart Disease and Diabetes datasets improves accuracy. Early stopping is based on achieving the best validation accuracy.}
\label{tab:classifier}
\end{table*}

\begin{table*}[t]
\centering
\renewcommand{\arraystretch}{1.3}
\setlength{\tabcolsep}{8pt}
\begin{tabular}{ll}
\hline
\textbf{Component}       & \textbf{Configuration} \\ \hline
\textbf{Discriminator}    & \begin{tabular}[t]{@{}l@{}}
\texttt{hidden\_dimensions} = [256, 256] \\
\texttt{pac\_dim} = 10 \\
\texttt{discriminator\_steps} = 1 \\
\texttt{discriminator\_lr} = 2e-4 \\
\texttt{discriminator\_decay} = 1e-6 \\
\texttt{adam\_betas} = [0.5, 0.9]
\end{tabular} \\ \hline
\textbf{Generator}        & \begin{tabular}[t]{@{}l@{}}
\texttt{hidden\_dimensions} = [256, 256] \\
\texttt{generator\_lr} = 2e-4 \\
\texttt{generator\_decay} = 1e-6 \\
\texttt{adam\_betas} = [0.5, 0.9]
\end{tabular} \\ \hline
\textbf{Other}            & \begin{tabular}[t]{@{}l@{}}
\texttt{batch\_size} = \{Adult/Employees: 500, Heart Disease/Diabetes: 5000\} \\
\texttt{epochs} = 100 (early stopping)
\end{tabular} \\ \hline
\end{tabular}
\caption{\textbf{Generator and Discriminator Configuration.} The batch size is increased tenfold for the Heart Disease and Diabetes datasets due to their larger size, to accelerate training. Early stopping is applied based on the highest fraction of counterfactuals on the validation set. Most model parameters are selected based on the CTGAN paper \cite{xu_modeling_2019}.}
\label{tab:ctgan}
\end{table*}

\begin{table*}[t]
\centering
\renewcommand{\arraystretch}{1.3}
\setlength{\tabcolsep}{12pt}
\begin{tabular}{ll}
\hline
\textbf{Parameter}                     & \textbf{Value} \\ \hline
\texttt{batch\_size}                   & Sample size (variable) \\
\texttt{classifier\_influence} ($\lambda_{\text{clas}}$) & 1 \\
\texttt{divergence\_influence} ($\lambda_{\text{div}}$) & 1 \\
\texttt{realism\_influence} ($\lambda_{\text{real}}$) & 0.1 \\
\texttt{adam\_betas}                   & [0.9, 0.999] \\
\texttt{lr}                            & 0.1 \\
\texttt{gradient\_steps}               & 30 \\ \hline
\end{tabular}
\caption{\textbf{Optimization Configuration.} Appendix Figure \ref{fig:optim finetune} shows that optimal performance is achieved within 30 gradient steps.}
\label{tab:config-optimization}
\end{table*}

\begin{figure*}[h]
    \centering
    \includegraphics[width=1\linewidth]{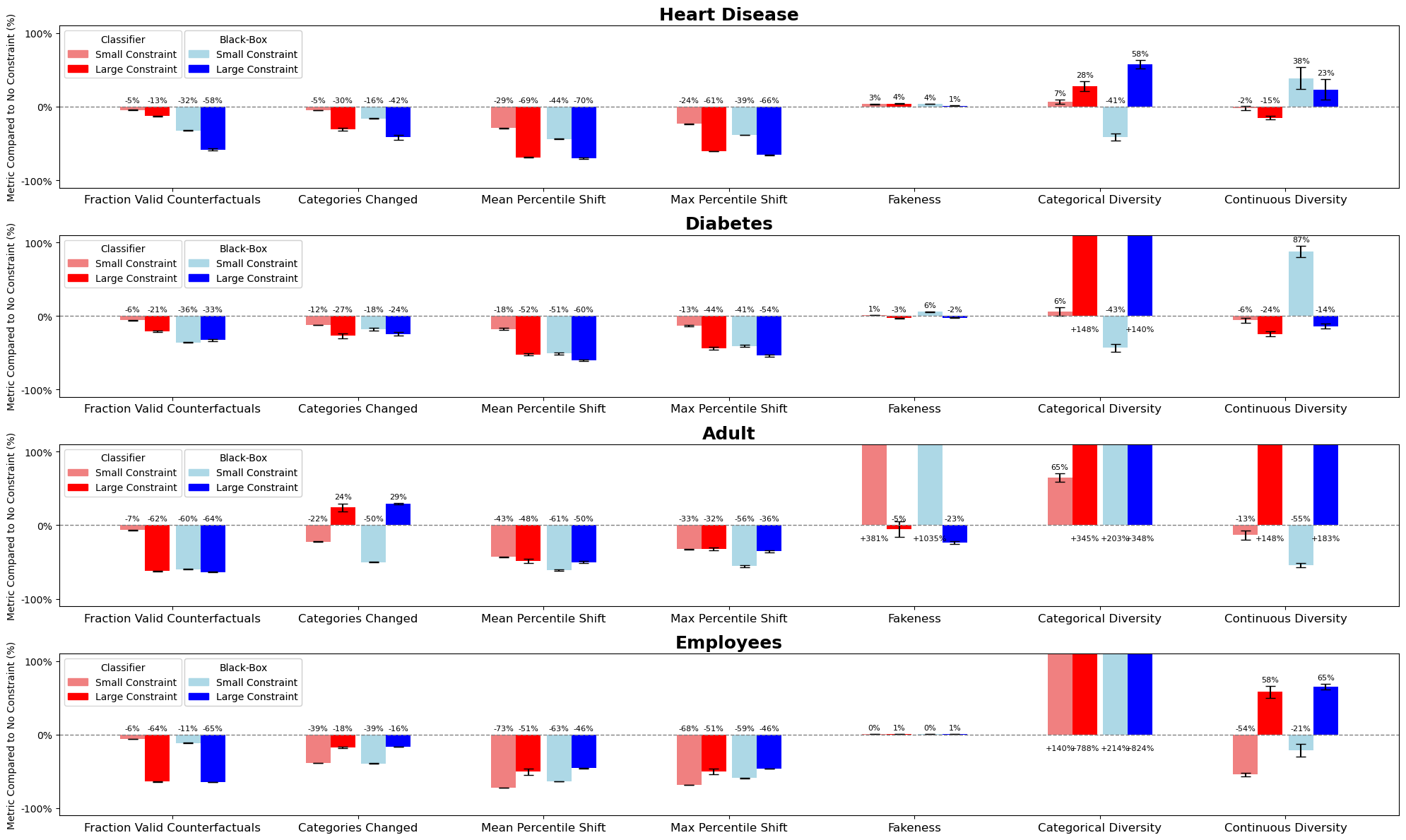}
    \caption{\textbf{Impact of Divergence Constraints on FCEGAN.} This figure illustrates the effect of applying a divergence constraint through a divergence loss (Eq. \ref{eq:div-loss}). Two levels of constraints, labeled as \textit{small} and \textit{large}, are evaluated to demonstrate the impact of increasingly stringent divergence constraints. Generally, the divergence constraint reduces divergence measures, affecting both categorical features (e.g., changes in categories) and continuous features (e.g., mean/max percentile shifts). Additionally, a divergence constraint can enhance diversity, which is decreased in template based methods, by mitigating mode collapse. The fraction of valid counterfactuals decreases as the divergence constraint becomes more stringent, reflecting the tighter constraints on the model. Notably, the level of fakeness remains largely unaffected across experiments, except for an anomaly observed in the Adult dataset under a small constraint. The presented measures represent the mean area under the curve (AUC) values of the flexibility graphs, as shown in Fig. \ref{fig:flexibility}, across five experiments. Error bars indicate the standard error of the mean (SEM). These values are normalized relative to implementations without divergence. Parameters used: $\lambda_{clas} = (1/0)$, $\lambda_{D_{og}} = 0.5$, $\lambda_{D_{cf}} = 0.5$. Mutable divergence influence ($\lambda_{m}$) values per dataset were as follows: small constraint with classifier: 10; large constraint with classifier: 100; small constraint in black-box setting: 5; and large constraint in black-box setting: 50.}
    \label{fig:divergence-impact}
\end{figure*}

\clearpage

\begin{figure*}[t]
     \centering
\includegraphics[width=1\linewidth]{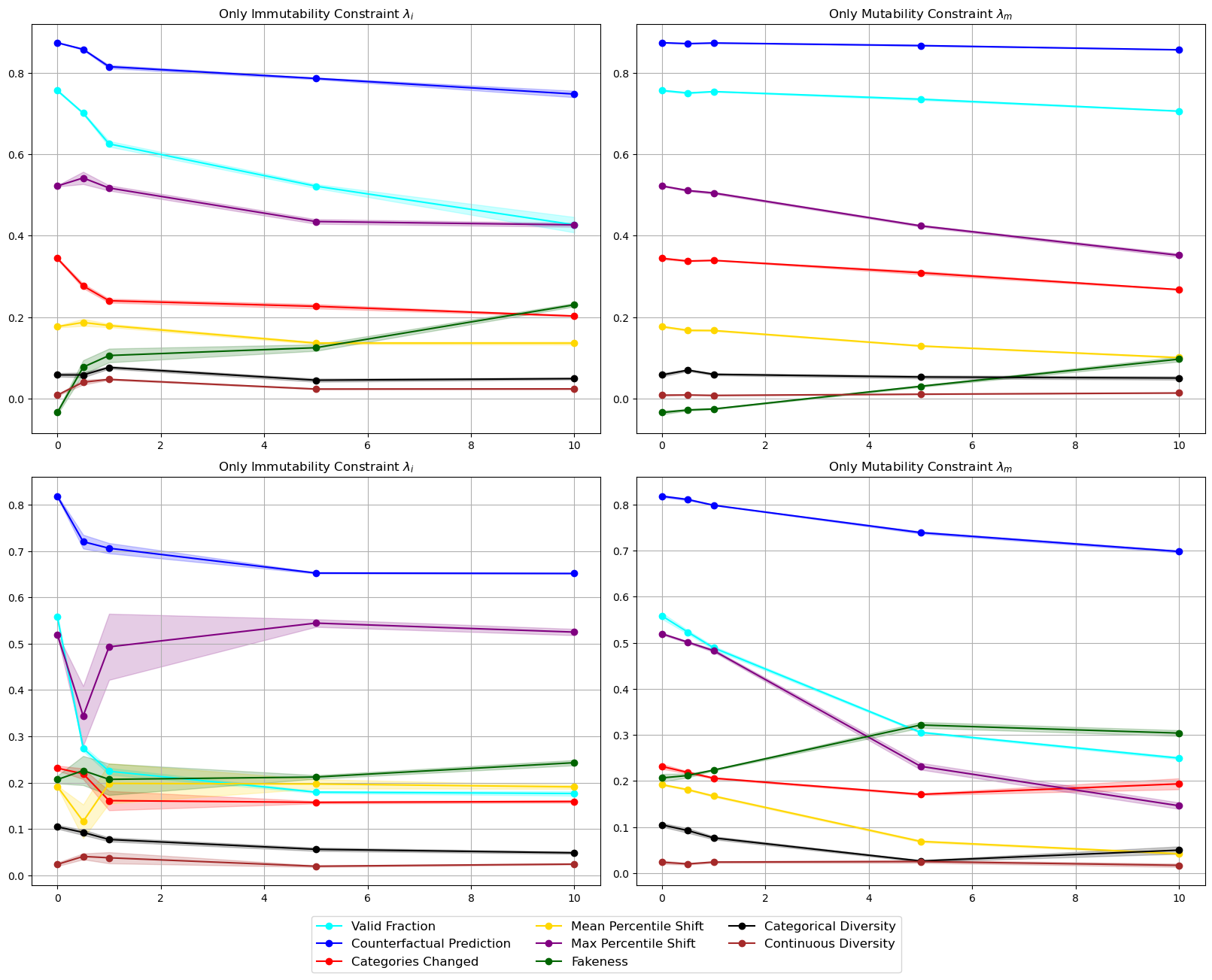}
\caption{\textbf{Hyperparameter Fine Tuning Divergence Constraints in FCEGAN (Adult).} In our implementation, we impose two divergence constraints: (1) an \textbf{Immutability Constraint} (regulated by $\lambda_i$) and (2) a \textbf{Mutability Constraint} (regulated by $\lambda_m$). The impact of $\lambda_i$ is diminished because immutable features are eventually reset, making its effect on reducing divergence measures weaker. In contrast, the \textbf{Mutability Constraint} ($\lambda_m$) more effectively reduces divergence, particularly for continuous divergence, while also maintaining a higher valid fraction compared to constraining immutable features ($\lambda_i$). Given its stronger impact, we primarily focus on the \textbf{Mutability Constraint} throughout this paper. The presented measures represent the mean area under the curve (AUC) values of the flexibility graphs, as shown in Fig. \ref{fig:flexibility}, across five experiments. Error bars indicate the standard error of the mean (SEM). All experiments are conducted with the following parameters: $\lambda_{clas}=(0/1)$; $\lambda_{D_{og}}=0.5$; $\lambda_{D_{cf}}=0.5$.}
     \label{fig:divergence-tune}
 \end{figure*}

\begin{figure*}[t]
    \centering
    \includegraphics[width=0.7\linewidth]{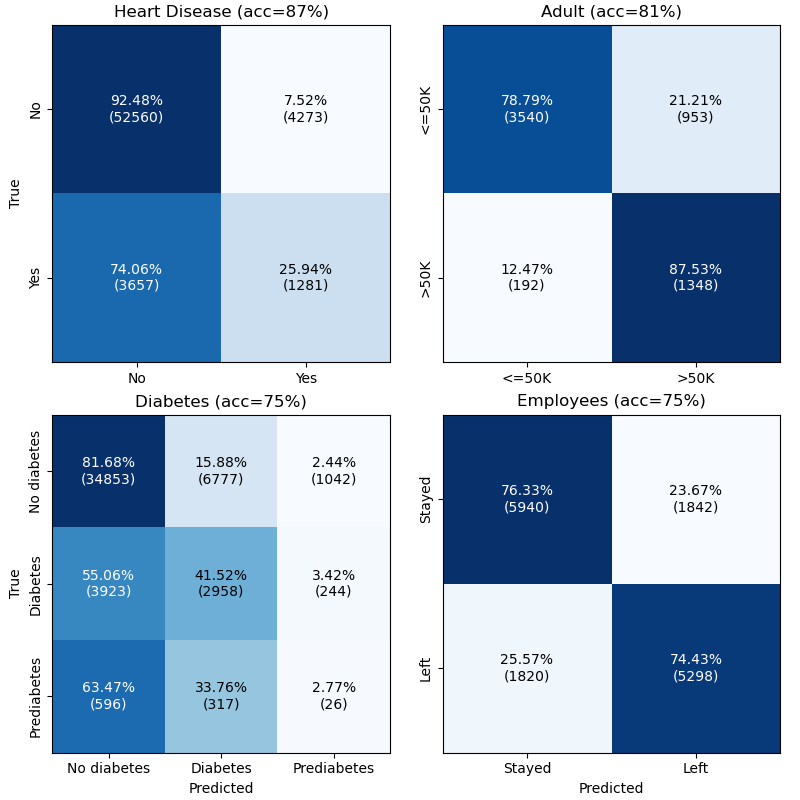}
    \caption{\textbf{Confusion Matrices of Classifiers.} The weighted loss method effectively balances performance across unbalanced class labels in the Adult and Employees datasets. However, it fails to achieve similar results for the Heart Disease Risk and Diabetes datasets, where significant imbalance remains in label predictions. This imbalance will further propagate to the dataset used for counterfactual explanations and should be taken into account. Performance is evaluated on the test dataset.}
    \label{fig:confusion classifier}
\end{figure*}

\begin{figure*}[t]
    \centering
    \includegraphics[width=0.85\linewidth]{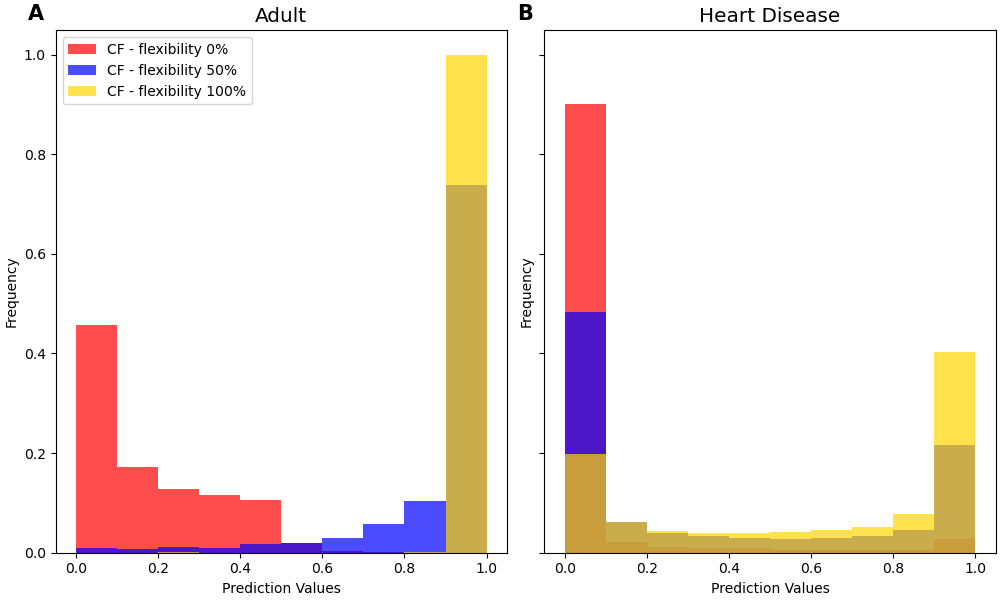}
   \caption{\textbf{Distributions of Counterfactual Predictions with Varying Flexibilities.} The counterfactual predictions increase with greater flexibility. The user has the agency to choose which features to set as mutable. Lower flexibility results in sparser and more actionable counterfactuals, but a lower fraction of valid counterfactuals. Additionally, depending on the use case, the threshold for a valid counterfactual explanation can be made more stringent. A Classifier FCEGAN (Adult/Heart Disease) is used with the following parameters respectively: $\lambda_{clas}=1$; $\lambda_{D_{og}}=0.5$; $\lambda_{D_{cf}}=0.5$; immutable divergence influence $\lambda_{i}= (1/10)$. The CTGAN data transformation is used, employing GMM-based normalization for continuous features.}
    \label{fig:dist counterfactual preds}
\end{figure*}

\begin{figure*}[t]
    \centering
    \includegraphics[width=0.9\linewidth]{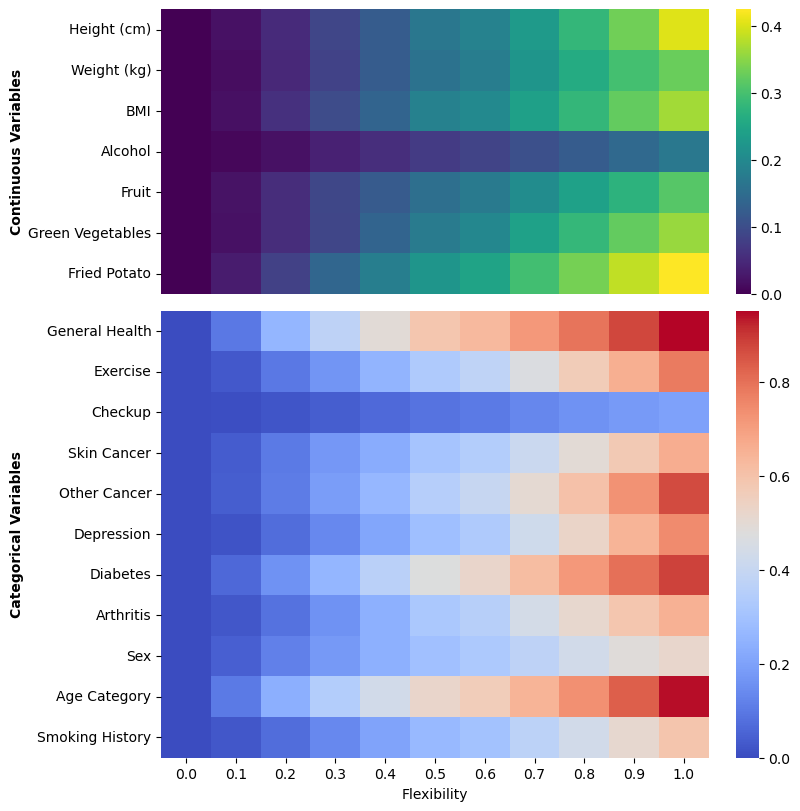}
   \caption{\textbf{Feature Divergence Measures in Counterfactual Explanation Generation (Heart Disease).} Average divergence measures (Table \ref{tab:measures}) are calculated for valid counterfactuals with different flexibilities. Divergence, as expected, increases with higher flexibility. However, the amount of divergence varies significantly across different features. Greater divergence indicates greater importance of that feature, which can be utilized in a fairness setting as well. A Classifier FCEGAN is used with the following parameters: $\lambda_{clas}=1$; $\lambda_{D_{og}}=0.5$; $\lambda_{D_{cf}}=0.5$; no divergence influences $(\lambda_i, \lambda_m)=(0,0)$. The CTGAN data transformation is used, employing GMM-based normalization for continuous features.}
    \label{fig:divergence heatmap}
\end{figure*}

\begin{figure*}[t]
    \centering
    \includegraphics[width=1\linewidth]{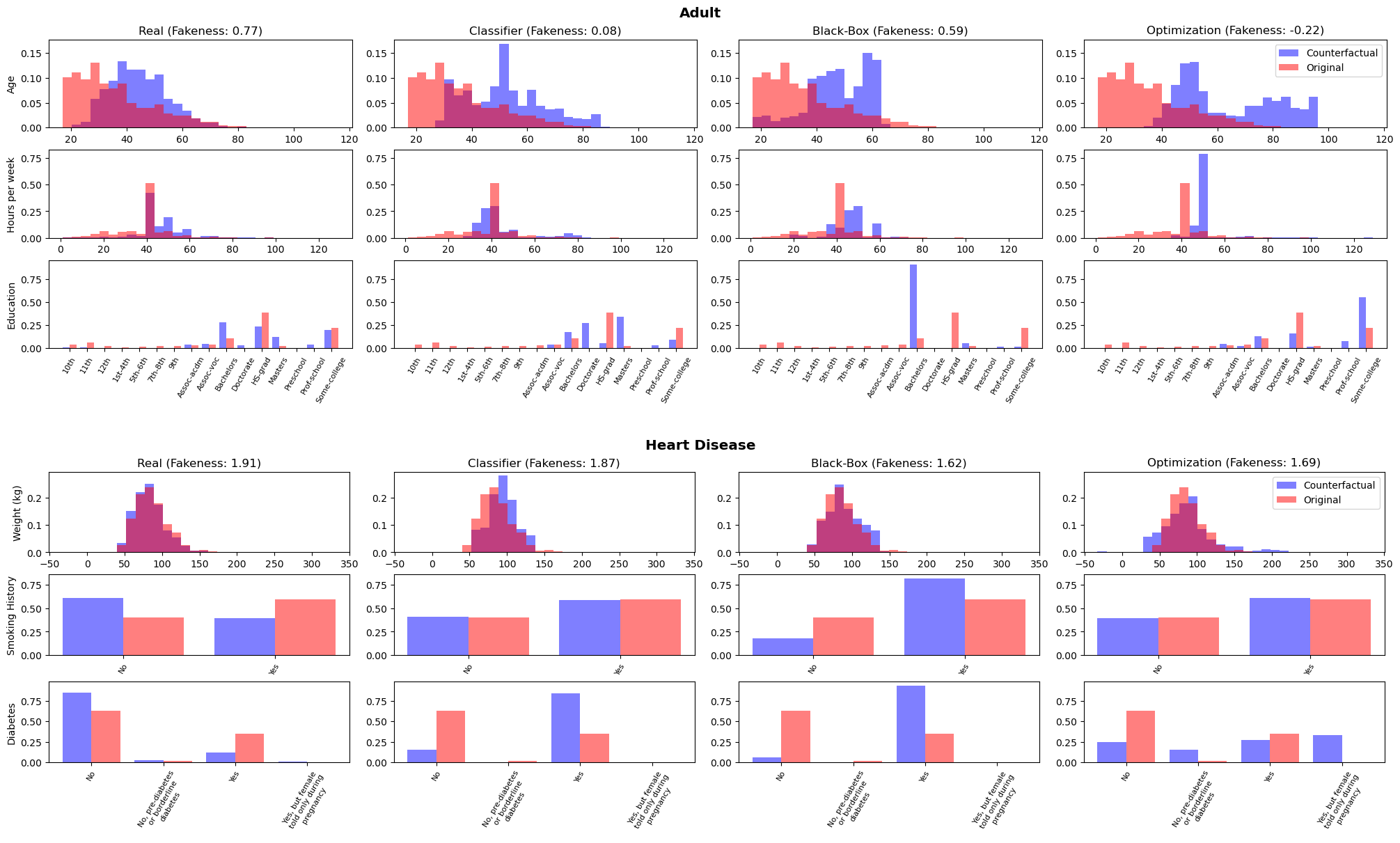}
   \caption{\textbf{Valid Counterfactual Explanation Distributions with 100\% Flexibility.} Counterfactuals are searched for all samples with an undesired predicted class label (``$\leq$ 50K wage'' / 'heart disease present') in the test dataset. Counterfactual templates are set to 100\% flexibility, allowing comparison with real counterfactuals (``$>$ 50K wage'' / 'no heart disease') present in the test dataset. The features shown are thus a subset of all the changed ones. Differences in realism and diversity between the distributions can be observed. The fakeness measure, obtained from an independent CTGAN discriminator (different from the one used during optimization), does not replace manual inspection of realism. Real data should have the lowest fakeness. The Age distribution further indicates that the Classifier FCEGAN and Optimization methods generate less realistic instances, with an overrepresentation of older ages, despite showing lower fakeness scores than real data. The following parameters are used for the (Adult+Classifier/Heart Disease+Classifier/Adult+Black-Box/Heart Disease+Black-Box) FCEGAN respectively: $\lambda_{clas}= (1/1/0/0)$;  immutable divergence influence $\lambda_{i}= (1/10/0.5/5)$; $\lambda_{D_{og}}=0.5$; $\lambda_{D_{cf}}=0.5$. The CTGAN data transformation is used, employing GMM-based normalization for continuous features.}

    \label{fig:whole dist}
\end{figure*}

\begin{figure*}[t]
    \centering
    \includegraphics[width=1\linewidth]{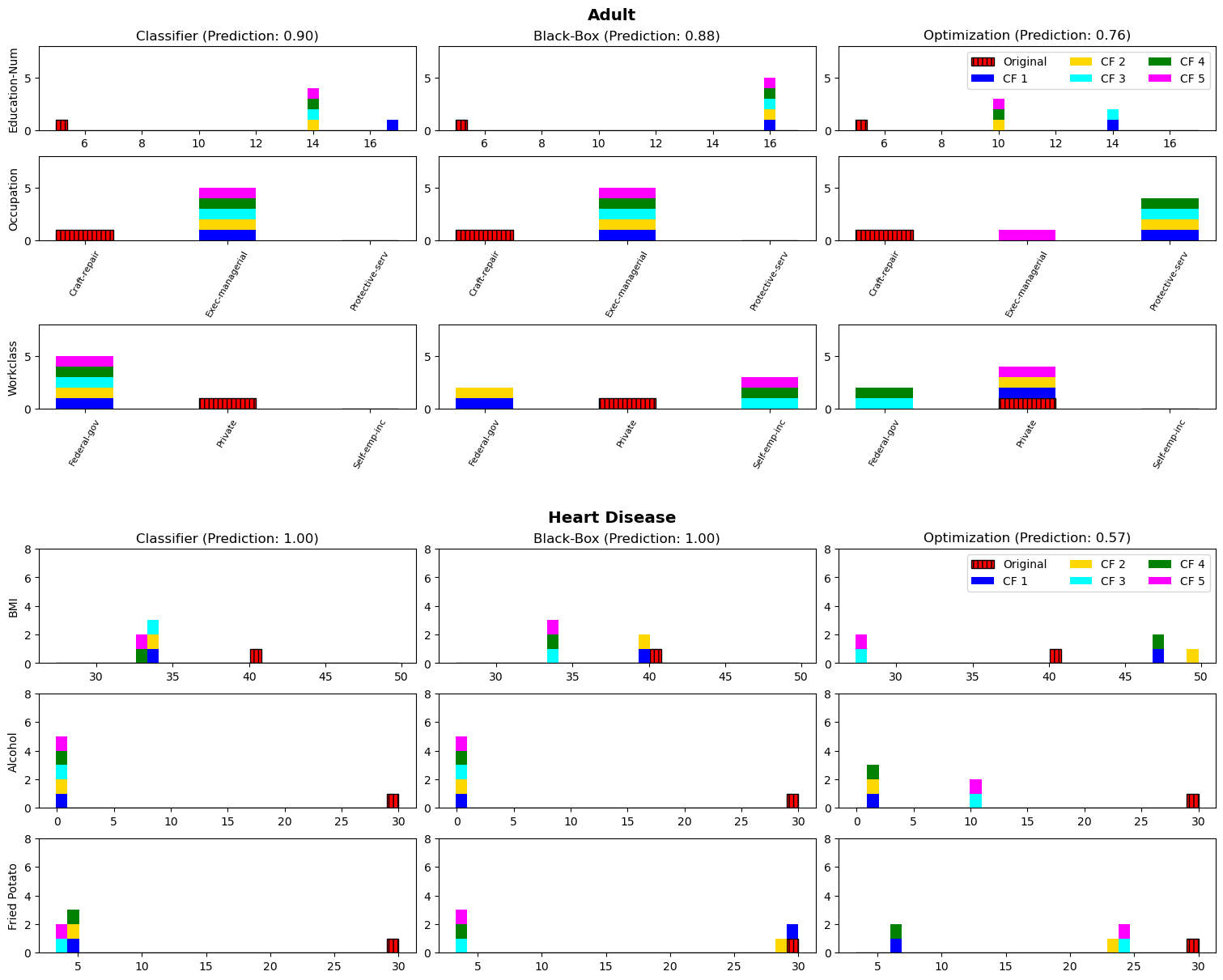}
   \caption{\textbf{Valid Counterfactual Explanation Examples.} For both datasets, three features are designated as mutable in the counterfactual template for a random sample with an undesired predicted class label (`$\leq$ 50K wage' / `heart disease present'). All other features are set as immutable. The task of generating five counterfactuals is assigned to three methods: Classifier FCEGAN, Black-Box FCEGAN, and a template-based optimization. The average prediction value of the counterfactuals is recorded for each method.  The following parameters are used for the (Adult+Classifier/Heart Disease+Classifier/Adult+Black-Box/Heart Disease+Black-Box) FCEGAN respectively: $\lambda_{clas}= (1/1/0/0)$; immutable divergence influence $\lambda_{i}= (1/10/0.5/5)$; $\lambda_{D_{og}}=0.5$; $\lambda_{D_{cf}}=0.5$. The CTGAN data transformation is used, employing GMM-based normalization for continuous features.}
    \label{fig:one sample dist}
\end{figure*}

\end{document}